\pdfoutput=1

\documentclass[11pt]{article}

\usepackage[]{acl}

\usepackage{times}
\usepackage{caption}
\usepackage{subcaption}
\usepackage{graphicx}
\usepackage{latexsym}
\usepackage{booktabs}
\usepackage{xcolor,colortbl}
\usepackage{amsmath}
\usepackage{multirow}
\usepackage[normalem]{ulem}
\usepackage{lipsum}
\usepackage{footmisc}

\usepackage{amssymb}
\usepackage{pifont}
\usepackage[T1]{fontenc}

\usepackage[utf8]{inputenc}

\usepackage{microtype}
\usepackage{xurl}
\usepackage{pdflscape}
\usepackage{geometry}
\usepackage{longtable}
\usepackage{enumitem}

\usepackage{inconsolata}
\usepackage{xcolor}

\title{PICK: Polished \& Informed Candidate Scoring \\
for Knowledge-Grounded Dialogue Systems}

\author{
Bryan Wilie, Yan Xu, Willy Chung, \\
\textbf{Samuel Cahyawijaya, Holy Lovenia, Pascale Fung}\\
Center for Artificial Intelligence Research (CAiRE)\\
The Hong Kong University of Science and Technology\\
\texttt{bwilie@connect.ust.hk}
}

\begin{document}
\maketitle
\begin{abstract}


Grounding dialogue response generation on external knowledge is proposed to produce informative and engaging responses. However, current knowledge-grounded dialogue (KGD) systems often fail to align the generated responses with human-preferred qualities due to several issues like hallucination and the lack of coherence.
Upon analyzing multiple language model generations, we observe the presence of alternative generated responses within a single decoding process. These alternative responses are more faithful and exhibit a comparable or higher level of relevance to prior conversational turns compared to the optimal responses prioritized by the decoding processes.
To address these challenges and driven by these observations, we propose Polished \& Informed Candidate Scoring (PICK), a generation re-scoring framework that empowers models to generate faithful and relevant responses without requiring additional labeled data or model tuning. Through comprehensive automatic and human evaluations, we demonstrate the effectiveness of PICK in generating responses that are more faithful while keeping them relevant to the dialogue history.
Furthermore, PICK consistently improves the system's performance with both oracle and retrieved knowledge in all decoding strategies. We provide the detailed implementation in \footnote{\url{https://github.com/bryanwilie/pick}}.

\end{abstract}


\section{Introduction}

\begin{figure*}
\centering
  \includegraphics[width=\textwidth]{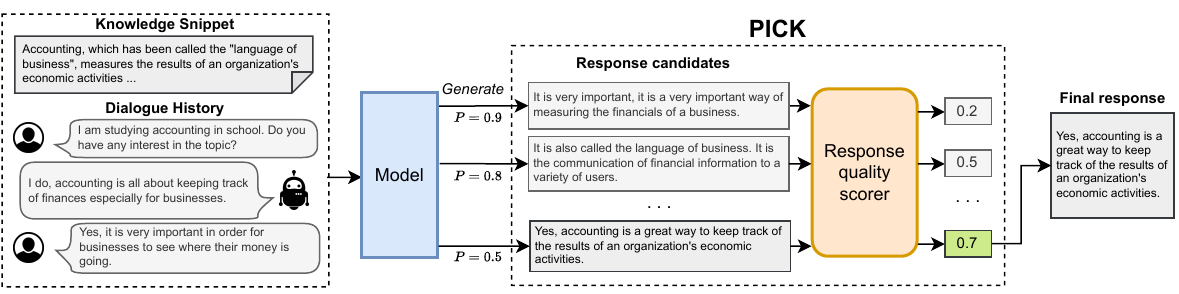}
  \caption{Overview of PICK. Instead of taking the response with the highest joint probability over the generated tokens, we select the response with the highest overall response quality score on faithfulness and relevance from the top-$r$ responses. We propose to assess the response candidates' quality based on the dialogue history and the corresponding knowledge without further tuning. Simple yet effective, PICK ensures better relevance and coherence of the generated response.
}
  \label{fig:roughmethod}
\end{figure*}

\begin{table}
\centering
\resizebox{\linewidth}{!}{%
\begin{tabular}{p{0.125\linewidth} p{0.875\linewidth}}
\toprule
\multicolumn{2}{l}{\textbf{Knowledge snippet}} \\
\multicolumn{2}{l}{Due to his powerful and very large vocal range and energetic live} \\
\multicolumn{2}{l}{performances, Rose has been \colorbox[HTML]{85c3ec}{named one of the greatest singers}} \\
\multicolumn{2}{l}{\colorbox[HTML]{85c3ec}{of all time by various media outlets}, including "Rolling Stone"} \\
\multicolumn{2}{l}{and "NME".} \\ \midrule
\multicolumn{2}{l}{\textbf{Dialogue History}} \\
\multicolumn{2}{l}{Speaker 1: Didn't their guitarist slash leave the band?} \\
\multicolumn{2}{l}{Speaker 2: When did he leave did he release the six albums with them} \\
\multicolumn{2}{l}{Speaker 1: I heard Axl Rose \colorbox[HTML]{ff5348}{was known for throwing tantrums}} \\ \midrule
\multicolumn{2}{l}{\textbf{Response}} \\
Vanilla & \multicolumn{1}{l}{He was \colorbox[HTML]{ff5348}{known for throwing tantrums}} \\
\textbf{PICK} & \multicolumn{1}{l}{He has been \colorbox[HTML]{85c3ec}{named one of the greatest singers of all time by}} \\
 & \multicolumn{1}{l}{\colorbox[HTML]{85c3ec}{various media outlets}} \\
\bottomrule
\end{tabular}%
}
\caption{PICK empowers models to generate more faithful to the knowledge snippet and serve as a more appropriate reply to the conversational context. In this sample, the response prioritized by PICK is more grounded in external knowledge (highlighted in blue) than the optimal response prioritized by the decoding process (i.e., beam search) and is also more relevant to the dialogue context. Here, the vanilla response repeats the dialogue history (highlighted in red).}
\label{tab:illustration}
\end{table}

Knowledge-grounded dialogue (KGD) has been introduced as a means to ground conversation towards the provided knowledge, thereby enabling the generation of informative and engaging responses~\cite{dinan2019wizard, zhou2018commonsense}. Despite the advancements in training KGD systems to convincingly simulate human language on a linguistic plane, these systems still struggle with the challenge of producing responses that align with those human-preferred qualities. Such deficits can be attributed to various issues, e.g., hallucination as well as the lack of coherence and engagingness in the generated responses~\cite{fu2022there, shuster2022language, rashkin2021increasing, zhao2020low}.

Numerous methodologies have been investigated to leverage the potential of various training and decoding methods to address these identified issues. For instance, the recent human quality alignment methods, such as \citet{ouyang2022training}, hinge on collecting extensive human annotations, followed by the reward model fine-tuning to approximate human preference.
This process then guides the optimization of the language model (LM) through reinforcement learning. While this approach has demonstrated promising results, it is noteworthy that accumulating such a significant volume of manual human data is highly resource-intensive in terms of both time and human labor.


Through analyzing various LM generations, we observe that within one decoding process, there exist alternative generated responses that are more faithful and relevant to prior conversational turns. These candidates, however, are overlooked by the decoding processes as they are not prioritized as the optimal responses.
Driven by these observations, we propose a straightforward yet effective human-aligned re-ranking framework to direct model responses closer to KGD qualities.

We introduce Polished \& Informed Candidate Scoring (PICK), a generation re-scoring framework for KGD tasks, which empowers models to generate optimal
dialogue responses that are more faithful to the knowledge provided and relevant to the dialogue history without requiring additional model tuning. The proposed framework is also model-agnostic; thus, it can be applied to various LMs with different architectures and sizes. Furthermore, it circumvents the need for supplementary labeled data by exploiting off-the-shelves metrics that correlate well with human judgment. While considering its contextual relevance to the dialogue history, utilizing these metrics allows the model to produce better responses.
However, to enable the generation of responses that are more faithful and relevant,
it is essential to condition the response on the dialogue history and accurate knowledge grounding. To do so, we explore various metrics that ensure the response is aligned with the knowledge and utilize the existing automatic metrics that correlate well with human judgment.
Our experiments and human evaluation show that PICK enables models to produce responses more faithful to the provided knowledge and relevant to the dialogue history. 

Our contributions to this work are three-fold. (1) We propose PICK, a generation re-scoring framework for KGD that empowers models to generate dialogue responses that are more faithful to the provided knowledge and relevant to the dialogue history. The proposed framework is simple yet effective; it does not require further model tuning and additional labelled data for language modelling alignment. (2) We analyze the improvement from PICK-reranked responses in the systems with both oracle and retrieved knowledge and show that PICK consistently improves the performance in all decoding strategies. (3) We investigate the impact of diverse scoring metrics and decoding settings on generation quality. Then, we present the best scoring and decoding configurations for PICK on KGD tasks.


\section{Related Work}


\paragraph{Knowledge-Grounded Dialogue} 
\citet{dinan2019wizard} develop a large dataset with conversations directly grounded on knowledge retrieved from Wikipedia. Alongside the work, recent works aim to build dialogue models that could conduct faithful and relevant knowledgeable discussions on open-domain topics~\cite{li2022knowledge, liu2021three, xu2022retrieval, xu2023kilm}. Aiming to improve informativeness, a knowledge selection process is introduced to determine which specific elements of knowledge are informative to the dialogue \cite{kim2020sequential, zhao2020knowledge}. Further, \citet{li2020zero} propose learning how knowledge is expressed to improve coherence and knowledge relevance. \citet{shuster2021retrieval} utilize neural-retrieval-in-the-loop architectures to develop models that maximize knowledgeability while retaining conversational ability. \citet{rashkin2021increasing} use the gold knowledge and control the model to generate faithful and relevant responses. The PICK framework is orthogonal to recent works but similarly focused on enabling the model to generate relevant responses faithful to the provided knowledge.


\paragraph{Alignment of Dialogue Response Quality}
To exhibit alignment with dialogue response quality in the model dialogue response, several works implemented the concept of reinforcement learning from human feedback~\cite{christiano2017deep} explicitly or implicitly in the dialogue domain. \citet{jaques2019way} evaluate responses for coherence and engagement using a supervised conversational evaluator with human-annotated labels. \citet{yi2019towards} collect human interaction data as implicit human feedback. \citet{hancock2019learning} develop an agent that would ask for feedback to improve its dialogue abilities further. Those works accumulate manual human data and are highly resource-intensive in terms of time and human labor. On the other hand, there are also works that re-rank response candidates to improve the dialogue response quality. \citet{mei2017coherent} utilize the Latent Dirichlet Allocation (LDA) method to learn document-level latent topics to select the best continuation based on document-level topic-level matching. \citet{welleck2018dialogue} improve the consistency by re-rank utterances using an NLI model trained on a Dialogue NLI dataset that they created for the purpose. \citet{ko2019linguistically} train four classifiers on synthetically generated data to re-rank plausible sentences. Unlike those works, we leverage off-the-shelf automatic metrics that correlate well with human judgment on conversation-level qualities; hence it does not require additional labeled data for the language modeling alignment. Furthermore, we devise a framework that doesn't require further model tuning to promote faithful and relevant candidates.

\section{Methodology}
\label{sec:method}

Knowledge-grounded dialogue (KGD) systems are built to be informative teachers. Such systems must be faithful to one or more source documents we implicitly trust and serve as an appropriate reply to the conversational context \cite{Rashkin2021measuring, zhan2021augmenting, honovich2021q2}. In KGD systems, a model is trained to generate a response based on the dialogue utterances with the user and ground to the knowledge snippet. We denote a KGD dataset as $\{\mathcal{D}^n\}^N_{n=1}$. At every turn $t$ we have dialogue history at turn $t$ denoted as $\mathcal{D}_t = \{(U_i,S_i)\}^t_{i=1}$, where $U_t$ is the user utterance and $S_t$ the system response. Each of these $S_t$ responses is grounded to knowledge snippets $K_t$ that are retrieved from a knowledge base
As illustrated in Figure \ref{fig:roughmethod}, our proposed framework takes in input $X_t = (T_t, K_t, \mathcal{D}_{t-1}, U_t)$ with $T_t$ resembling the conversation topic at turn t, to a fine-tuned model $f_\theta$ to generate a relevant and faithful response sequence $\hat{S}_t$. The concatenation of $\mathcal{D}_{t-1}$ and $U_t$ is a dialogue history.

\subsection{Re-ranking Framework}
Beam search and nucleus sampling decoding methods allow the model to generate multiple responses (i.e., hypotheses) to the same inputs. Instead of selecting the response with the highest probability from the model, we propose a re-ranking method that ensures better relevance and faithfulness of the generated responses without further tuning. Our approach treats all of the $r$ hypotheses as a pool of $r$ response candidates $C = \{C_1,..., C_r\}$ to be further ranked based on their qualities. We evaluate each response candidate with ready-to-use scorers to get its quality score of $\mu$. Our goal is to select the best scoring candidate according to their associated scores $C_{\mu}=\{\mu(C_1),..., \mu(C_r)\}$, that is, to identify the best dialogue response candidate $\hat{S}_t$ according to the metrics, which is given by:
\begin{align*}
    \hat{S}_t = \operatorname*{arg\,max}_{C_j \in C} \{\mu(C_1),..., \mu(C_r)\}
\end{align*}

\subsection{Decoding Strategy}
To produce the top-$r$ response candidates $C = \{C_1,..., C_r\}$, we take the same input $X_t$ to the fine-tuned model $f_\theta$ and perform generations with the number of return sequences to be $r$, with $r$ being larger than 1. Each response candidate $C_j$ is an independently computed returned sequence from the search hypotheses or random sampling. Although by both paradigms, the last $(r-1)$ hypotheses are seen as inferior response candidates, we will later show that it is not the case and that by evaluating their qualities using the ready-to-use automatic metric, we can let the same fine-tuned model $f_\theta$ reach a more optimal response quality.

\subsection{Response Quality Scorer}
\label{sec:scoring_framework}

KGD aims to ground the conversation by generating responses that are faithful to the provided knowledge and relevant to the dialogue history.
To achieve this, we leverage off-the-shelf automatic metrics to evaluate the quality of response candidates. These metrics allow us to assess the faithfulness and the relevance of the responses without the need for additional labeled data.
We consider the qualities of the response candidate w.r.t the dialogue history $\mu_D$ and the input knowledge snippet $\mu_K$ to construct the final quality score of $\mu$. We elaborate on the metric corresponding to each aspect score in Section \ref{sec:faithfulness_score} and \ref{sec:relevance_score}.

The relevance score $\mu_D$ is calculated given the response candidates and the dialogue history, $\mu_D(C_j, (\mathcal{D}_{t-1}, U_t))$, which the faithfulness score $\mu_K$ is calculated regarding the input knowledge snippet, $\mu_K(C_j, K_t)$. In this work, we consider both qualities of the response candidate equally important; thus, we derive the final quality score $\mu$ based on the sum of $\mu_D$ and $\mu_K$.
Our proposed method allows more randomness in the decoding process, which may cause high meaningless repetition in some $r$ hypotheses. To filter this, we remove the hypotheses that contain repetitive words.
We also filter hypotheses that contain a word more than 30 characters long since words that long are not likely to occur in an English general text.\footnote{\url{https://en.wikipedia.org/wiki/Longest_word_in_English}}

\begin{table}[!t]
\centering
\resizebox{0.6\linewidth}{!}{%
\begin{tabular}{lcc}
\toprule
\multirow{2}{*}{\textbf{Split}} & \multicolumn{2}{c}{\textbf{\# Wizard responses}} \\ \cmidrule{2-3}
& \textbf{seen} & \textbf{unseen} \\ \midrule
Train & 74092 & - \\
Dev & 3939 & 3927 \\
Test  & 3865 & 3924 \\ \bottomrule
\end{tabular}%
}
\caption{Statistics of Wizard of Wikipedia.}
\label{tab:datasetstats}
\end{table}

\section{Experiments}
\subsection{Dataset and Models}

We use Wizard of Wikipedia (WoW)~\cite{dinan2019wizard}, a large-scale corpus of multi-turn knowledge-grounded dialogues between an ``apprentice'' and a ``wizard'', to conduct our experiments in developing the KGD systems. We use the same split in \cite{dinan2019wizard} as stated in \cite{shuster2020dialogue}. We aim to produce better responses, thus we focus on only modeling ``wizard'' response utterances in the dialogue where they respond to the ``apprentice'' utterances. The data statistics of WoW are shown in Table~\ref{tab:datasetstats}. We adopt a pre-trained GPT-2 \cite{radford2019language}, and T5-small \cite{raffel2020exploring} as the backbones. We fine-tune both models and limit the maximum sequence length to 512. Maximizing our GPU (RTX 2080Ti) capacity, we train the GPT-2 model in a batch size of 4 and the T5 model in a batch size of 8 for 10 epochs with early stopping patience of 3. We train using all of the training data and use the dev (seen topics) split and monitor the model's loss in this split to do the early stopping and to choose the best model to use in the experiment. More training details are provided in \S\ref{sec:training_details}.



\subsection{Faithfulness Score}
\label{sec:faithfulness_score}

Faithfulness problem can also be considered an intrinsic hallucination problem for KGD tasks. Following \citet{shuster2021retrieval}, we leverage Knowledge F1 (KF1)\footnote{\label{f1parlai}\url{https://github.com/facebookresearch/ParlAI}}, calculated based on the unigram overlap between the generated response and the input knowledge snippet, to assess the faithfulness of responses. There are also other alternative n-gram-based automatic metrics such as BLEU~\cite{papineni2002bleu}, ROUGE~\cite{Lin2004ROUGEAP}, entailment measurement from a state-of-the-art natural language interference (NLI) model~\cite{liu2019roberta}, and the similarity measurement from BLEURT~\cite{Sellam2020BLEURTLR}. We further investigate the impact of different faithfulness scorers in Section~\ref{sec:diff_metrics} and find out that KF1 shows its distinct effectiveness in ensuring both faithfulness and overall performance.


\begin{table*}[!t]
\centering
\resizebox{\textwidth}{!}{%
\begin{tabular}{lcccccccc}
\toprule
 & \multicolumn{4}{c}{\textbf{Test (seen topics)}} & \multicolumn{4}{c}{\cellcolor[HTML]{FFFFFF}\textbf{Test (unseen topics)}} \\ \cmidrule(lr){2-5} \cmidrule(lr){6-9}
\multirow{-2}{*}{\textbf{Models}} & \textbf{BLEU-4} & \textbf{ROUGE-L} & \textbf{F1} & \textbf{KF1} & \textbf{BLEU-4} & \textbf{ROUGE-L} & \textbf{F1} & \textbf{KF1} \\ \midrule
\rowcolor[HTML]{C0C0C0} 
\textbf{Baselines} & \multicolumn{1}{l}{\cellcolor[HTML]{C0C0C0}} & \multicolumn{1}{l}{\cellcolor[HTML]{C0C0C0}} & \multicolumn{1}{l}{\cellcolor[HTML]{C0C0C0}} & \multicolumn{1}{l}{\cellcolor[HTML]{C0C0C0}} & \multicolumn{1}{l}{\cellcolor[HTML]{C0C0C0}} & \multicolumn{1}{l}{\cellcolor[HTML]{C0C0C0}} & \multicolumn{1}{l}{\cellcolor[HTML]{C0C0C0}} & \multicolumn{1}{l}{\cellcolor[HTML]{C0C0C0}} \\ \midrule
MemNet (w/ aux loss)~\cite{dinan2019wizard} & 1.5 & - & 35.5\% & - & 0.3 & - & 32.2\% & - \\
dodecaDialogue~\cite{shuster2020dialogue} & 10 & - & \underline{\textbf{38.4\%}} & - & \textbf{9.7} & - & - & - \\
Controlled GPT-2~\cite{rashkin2021increasing} & 8.9 & - & - & - & 8.4 & - & - & - \\
Controlled T5~\cite{rashkin2021increasing} & 8.4 & - & - & - & 8.7 & - & - & - \\ 
PLUG-Golden Knowledge~\cite{li2022knowledge}  & \textbf{11.5} & \textbf{31.1} & 36.0\% & \textbf{47.8\%} & 8.8 & \textbf{29.0} & \textbf{33.4\%} & \textbf{46.0\%} \\ \midrule
\rowcolor[HTML]{C0C0C0} 
{\color[HTML]{333333} \textbf{GPT-2}} & \multicolumn{1}{l}{\cellcolor[HTML]{C0C0C0}{\color[HTML]{333333} }} & \multicolumn{1}{l}{\cellcolor[HTML]{C0C0C0}{\color[HTML]{333333} }} & \multicolumn{1}{l}{\cellcolor[HTML]{C0C0C0}{\color[HTML]{333333} }} & \multicolumn{1}{l}{\cellcolor[HTML]{C0C0C0}{\color[HTML]{333333} }} & \multicolumn{1}{l}{\cellcolor[HTML]{C0C0C0}{\color[HTML]{333333} }} & \multicolumn{1}{l}{\cellcolor[HTML]{C0C0C0}{\color[HTML]{333333} }} & \multicolumn{1}{l}{\cellcolor[HTML]{C0C0C0}{\color[HTML]{333333} }} & \multicolumn{1}{l}{\cellcolor[HTML]{C0C0C0}{\color[HTML]{333333} }} \\ \midrule
Greedy & 12.4 & 29.9 & 32.6\% & 48.6\% & 12.1 & 29.9 & 32.3\% & 46.8\% \\ \midrule
Beam search ($n$ = 5, $r$ = 1) & 15.0 & 33.1 & 35.6\% & 64.5\% & 13.9 & 32.2 & 34.5\% & 60.3\% \\
\quad + PICK ($n$ = 5, $r$ = 5) & \textbf{16.6} & \textbf{34.1} & \textbf{37.0\%} & \textbf{73.7\%} & \textbf{15.6} & \textbf{33.7} & \textbf{36.4\%} & \textbf{71.0\%} \\ \midrule
Beam search ($n$ = 10, $r$ = 1) & 15.4 & 33.4 & 35.8\% & 68.9\% & 14.3 & 32.5 & 34.7\% & 64.5\% \\
\quad + PICK ($n$ = 10, $r$ = 10) & \textbf{16.7} & \textbf{34.5} & \textbf{37.4\%} & \textbf{80.4\%} & \textbf{16.0} & \textbf{34.5} & \textbf{37.1\%} & \textbf{78.2\%} \\ \midrule
Top-$k$ sampling ($k$ = 3, $r$ = 1) & 8.7 & 26.3 & 29.0\% & 39.7\% & 8.1 & 25.5 & 28.2\% & 37.8\% \\
\quad + PICK ($k$ = 3, $r$ = 10) & \textbf{14.9} & \textbf{33.0} & \textbf{36.2\%} & \textbf{67.6\%} & \textbf{14.2} & \textbf{32.7} & \textbf{35.6\%} & \textbf{64.6\%} \\ \midrule
Top-$p$ sampling ($p$ = 0.5, $r$ = 1) & 11.5 & 28.3 & 31.2\% & 46.2\% & 10.4 & 27.6 & 30.1\% & 43.5\% \\
\quad + PICK ($p$ = 0.5, $r$ = 10) & \textbf{16.0} & \textbf{34.1} & \textbf{37.2\%} & \textbf{72.7\%} & \textbf{15.2} & \textbf{34.0} & \textbf{36.9\%} & \textbf{70.2\%} \\ \midrule
\rowcolor[HTML]{C0C0C0} 
\textbf{T5} & \multicolumn{1}{l}{\cellcolor[HTML]{C0C0C0}} & \multicolumn{1}{l}{\cellcolor[HTML]{C0C0C0}} & \multicolumn{1}{l}{\cellcolor[HTML]{C0C0C0}} & \multicolumn{1}{l}{\cellcolor[HTML]{C0C0C0}} & \multicolumn{1}{l}{\cellcolor[HTML]{C0C0C0}} & \multicolumn{1}{l}{\cellcolor[HTML]{C0C0C0}} & \multicolumn{1}{l}{\cellcolor[HTML]{C0C0C0}} & \multicolumn{1}{l}{\cellcolor[HTML]{C0C0C0}} \\ \midrule
Greedy & 14.7 & 33.0 & 35.6\% & 56.0\% & 14.4 & 32.4 & 35.0\% & 56.2\% \\ \midrule
Beam search ($n$ = 5, $r$ = 1) & 16.3 & 34.8 & 37.7\% & 77.8\% & 15.6 & 34.7 & 37.4\% & 78.8\% \\
\quad + PICK ($n$ = 5, $r$ = 5) & \textbf{16.3} & \underline{\textbf{34.9}} & \textbf{37.8\%} & \textbf{79.6\%} & \textbf{15.6} & \underline{\textbf{34.8}} & \underline{\textbf{37.6\%}} & \textbf{80.2\%} \\ \midrule
Beam search ($n$ = 10, $r$ = 1) & \textbf{16.2} & \textbf{34.8} & \multicolumn{1}{l}{37.7\%} & \multicolumn{1}{l}{81.8\%} & \textbf{15.5} & 34.7 & \multicolumn{1}{l}{37.5\%} & 82.7\% \\
\quad + PICK ($n$ = 10, $r$ = 10) & 16.1 & 34.8 & \textbf{37.7\%} & \underline{\textbf{84.3\%}} & 15.4 & \underline{\textbf{34.8}} & \underline{\textbf{37.6\%}} & \underline{\textbf{84.8\%}} \\ \midrule
Top-$k$ sampling ($k$ = 3, $r$ = 1) & 11.7 & 30.3 & 33.3\% & 49.3\% & 11.4 & 29.9 & 32.8\% & 49.8\% \\
\quad + PICK ($k$ = 3, $r$ = 10) & \textbf{15.9} & \textbf{34.2} & \textbf{37.3\%} & \textbf{71.8\%} & \textbf{15.2} & \textbf{34.0} & \textbf{37.0\%} & \textbf{72.4\%} \\ \midrule
Top-$p$ sampling ($p$ = 0.5, $r$ = 1) & 13.9 & 32.3 & 35.1\% & 55.3\% & 14.0 & 31.9 & 34.6\% & 55.6\% \\
\quad + PICK ($p$ = 0.5, $r$ = 10) & \underline{\textbf{16.9}} & \underline{\textbf{34.9}} & \textbf{38.0\%} & \textbf{74.6\%} & \underline{\textbf{16.4}} & \textbf{34.7} & \textbf{37.5\%} & \textbf{74.7\%} \\ \bottomrule
\end{tabular}%
}
\caption{Overall performance comparisons. PICK significantly improves the performances of all models and decoding methods, even on the top-$k$ and top-$p$ sampling that gained low automatic metrics scores on their vanilla responses.
The best performances in each section are in \textbf{bold}, while the overall best is \underline{underlined}.}
\label{tab:mainresults}
\end{table*}

\subsection{Relevance Score}
\label{sec:relevance_score}

Overlap-based automatic evaluation metrics are known to be ineffective in distinguishing the relevance between the generated response with the dialogue history due to the one-to-many nature of dialogue~\cite{zhao2017learning, yeh2021comprehensive}. 
Therefore, we explore reference-free model-based metrics on top of them. Specifically, we utilize the FED metric~\cite{mehri2020unsupervised} as the relevance scorer. FED is an unsupervised evaluation metric that uses DialoGPT \cite{zhang2020dialogpt} to measure 18 fine-grained turn- and dialogue-level qualities of dialogue. It calculates the likelihood of manually designed follow-up utterances to measure multiple qualities of dialogue. Moreover, it is proven to correlate well with human judgment.

We follow the hierarchical groupings from \cite{Phy2020DeconstructTR} and separate the fine-grained metrics in FED between basic (w.r.t understandability) and further (w.r.t likeability) response qualities, both at the turn- and dialogue-level. At the turn level (\textbf{TL}), we group semantically appropriate, understandable, and fluent as turn-level metrics that measure the basic qualities of responses. We see the additional qualities as the ones that make the response more likeable and group the interesting, engaging, specific, relevant, and correct measurements into one. Similarly, at the dialogue level (\textbf{DL}), we group coherent, error recovery, consistent, and diverse as the dialogue-level basic qualities of responses. At the same level, we also group depth, likeable, understandable, flexible, informative, and inquisitive as dialogue-level metrics that measure the further qualities of responses. On top of that, we also experiment with combining each level of metrics and all of the measurements to find the best combination to produce responses relevant to the dialogue history.

We also explore another reference-free model-based metric, USL-H~\cite{Phy2020DeconstructTR}, for comparison. USL-H combines three models trained to determine whether a response is valid and grammatically correct and to evaluate the sensibleness and the likelihood of a given response. The analysis is included in Section~\ref{sec:diff_metrics}.


\begin{table*}[t]
\centering
\resizebox{0.95\textwidth}{!}{%
\begin{tabular}{lcccccccc}
\toprule
\multirow{2}{*}{\textbf{Inference methods}} & \multicolumn{4}{c}{\textbf{Test (seen topics)}} & \multicolumn{4}{c}{\textbf{Test (unseen topics)}} \\ \cmidrule(lr){2-5} \cmidrule(lr){6-9}
 & \textbf{BLEU-4} & \textbf{ROUGE-L} & \textbf{F1} & \textbf{KF1} & \textbf{BLEU-4} & \textbf{ROUGE-L} & \textbf{F1} & \textbf{KF1} \\ \midrule
Greedy & \underline{5.8} & 20.3 & 22.0\% & 56.4\% & 4.7 & 19.0 & 20.4\% & 53.6\% \\ \midrule
Beam search ($n$ = 10, $r$ = 1) & 2.5 & 12.6 & 14.4\% & 47.3\% & 2.1 & 12.0 & 13.6\% & 45.6\% \\
\quad + PICK ($n$ = 10, $r$ = 10) & \textbf{2.5} & \textbf{12.8} & \textbf{14.7\%} & \textbf{47.9\%} & \textbf{2.1} & \textbf{12.1} & \textbf{13.9\%} & \textbf{46.1\%} \\ \midrule
Top-$k$ sampling ($k$ = 3, $r$ = 1) & 4.4 & 18.5 & 20.5\% & 46.7\% & 3.7 & 17.6 & 19.4\% & 45.3\% \\
\quad + PICK ($k$ = 3, $r$ = 10) & \textbf{5.4} & \textbf{19.5} & \textbf{21.3\%} & \underline{\textbf{70.4\%}} & \textbf{4.7} & \textbf{18.5} & \textbf{20.1\%} & \underline{\textbf{67.9\%}} \\ \midrule
Top-$p$ sampling ($p$ = 0.3, $r$ = 1) & 5.6 & 20.1 & 21.8\% & 56.0\% & 4.6 & 18.8 & 20.3\% & 52.9\% \\
\quad + PICK ($p$ = 0.3, $r$ = 10) & \underline{\textbf{5.8}} & \underline{\textbf{20.4}} & \underline{\textbf{22.1\%}} & \textbf{68.9\%} & \underline{\textbf{4.9}} & \underline{\textbf{19.3}} & \underline{\textbf{20.9\%}} & \textbf{66.2\%} \\ \bottomrule
\end{tabular}%
}
\caption{Results of KnowledGPT~\cite{zhao2020knowledge} with PICK and different inference methods. Although other decoding methods underperform the greedy baseline, our method still improves over each of them.
Here, KF1 is calculated w.r.t the retrieved knowledge instead of the gold knowledge. The best performances in each section are in \textbf{bold}, while the overall best is \underline{underlined}.}
\label{tab:knowledgpt}
\end{table*}

\subsection{Baselines}
We select baseline models that utilize
gold knowledge snippets in their generation process. We take the performances of MemNet \cite{dinan2019wizard}, dodecaDialogue \cite{shuster2020dialogue}, GPT-2 and T5 with control code and resampling \cite{rashkin2021increasing}, and PLUG-Golden Knowledge \cite{li2022knowledge} as our baselines. We also experiment with PICK in the settings where the provided knowledge is retrieved instead of using the oracle knowledge. We leverage KnowledGPT~\citet{zhao2020knowledge} and perform a similar procedure elaborated in Section \ref{sec:method}, to select responses that are more relevant to the dialogue history and the ones that are more faithful to the retrieved knowledge. We directly utilize the codes and models provided\footnote{\url{https://github.com/zhaoxlpku/KnowledGPT.git}} and adjust the generation parameters as required. 

We note the \textbf{vanilla} responses performance as our lower bound: the responses performance on each decoding method without PICK (i.e. taking the top-1 hypotheses from beam search). 

\subsection{Training details}
\label{sec:training_details}

During training, the concatenated utterances are delimited using speaker ID of either \texttt{<speaker1>} or \texttt{<speaker2>}, and the concatenations of the topic, knowledge snippet, and the utterances are separated by a separator token \texttt{\textbackslash n}. 
We experiment using learning rates (lr) of ${1e-5, 5e-4, 1e-4, 5e-4}$ to fine-tune the models, and then we pick the models with the lowest loss on the dev (seen topics) split to be the models that we will be using throughout the experiments. Ultimately, the best GPT-2 model is fine-tuned with lr of 1e-5 and the T5 with 5e-4. 

\subsection{Evaluation}

\paragraph{Automatic Metrics} We evaluate the final response qualities by comparing them to the gold responses. We perform the automatic evaluation using BLEU-4~\cite{papineni2002bleu}, ROUGE-L~\cite{Lin2004ROUGEAP}, and unigram-F1\footref{f1parlai}. We implement the BLEU measurements following~\citet{rashkin2021increasing}. To make a fair comparison with the previous work, we utilize BLEU-4 scoring as it is implemented in~\citet{rashkin2021increasing} and ROUGE-L scoring (the mean F1 measures) as it is implemented in~\footnote{\url{https://huggingface.co/spaces/evaluate-metric/rouge}}. Further, we also use KF1 as stated in Section~\ref{sec:faithfulness_score} for faithfulness measurement.


\paragraph{Human Evaluation} 
We conduct manual evaluations to measure the qualities of the generated responses from two aspects: \textit{Faithfulness} and \textit{Relevance}. We take 100 random generation samples from the test (unseen topics) split and ask crowd-sourced annotators~\footnote{\url{https://appen.com/}} to evaluate on a 4-point Likert scale from 1 (low quality) to 4 (high quality). We ask for three level-1 (all-kinds) contributors and three level-3 (experienced only) contributors and report their average scores.
The complete annotation guideline is attached in Appendix \ref{sec:appendix_annotation guidance}.

\section{Results}

\subsection{Results with Oracle Knowledge}
\label{sec:results_gold}

Overall, as observed in Table \ref{tab:mainresults}, the proposed method achieved significantly better performances than the baselines from the previous works, especially in the comparison of BLEU-4 scores. In this table, the PICK responses are all re-ranked based on the sum of FED turn-level basic metrics and KF1. The proposed method significantly improves the performances of all models and decoding methods in all of the BLEU-4, ROUGE-L, F1, and KF1 metrics. Interestingly, for the top-$k$ and top-$p$ sampling decoding that previously gained low scores BLEU-4, ROUGE-L, F1, and even KF1 on their vanilla responses, there exist alternative responses that have a more similar quality to the gold response and our proposed re-ranking and scoring framework promotes that.

All the re-ranked responses also obtained a huge increase of KF1 compared to their vanilla baseline, especially in the top-$k$ and the top-$p$ sampling, where the vanilla KF1 is far much lower than the re-ranked response's KF1. Although the scores of BLEU-4, ROUGE-L, and F1 are comparable in the inference of the T5 model using beam search, we can still see the improvement in KF1, signifying that the proposed method better addresses the use of knowledge in its response.
Lastly, PICK also closes the performance gap between unseen and seen evaluation, as it is especially observed in the GPT-2 model performances. We provide samples of the responses in Table \ref{tab:cherrypicked_samples}.

\begin{table}[t]
\centering
\resizebox{0.95\columnwidth}{!}{%
\begin{tabular}{l|ll}
\toprule
\textbf{Models} & \textbf{Faithfulness} & \textbf{Relevance}\\ \midrule
Gold responses & 2.64 & 2.09\\ \midrule
\multicolumn{3}{l}{\cellcolor[HTML]{C0C0C0}\textbf{GPT-2}} \\
Beam search ($n$ = 10, $r$ = 1) & 2.26 & 1.48\\
 + PICK ($n$ = 10, $r$ = 10) & \textbf{2.49*} & \textbf{1.86*} \\ \midrule
\multicolumn{3}{l}{\cellcolor[HTML]{C0C0C0}\textbf{T5}} \\
Beam search ($n$ = 10, $r$ = 1) & 2.51 & 1.87\\
 + PICK ($n$ = 10, $r$ = 10) & \textbf{2.56} & \textbf{2.01*}\\ \bottomrule
\end{tabular}%
}
\caption{PICK enables models to produce responses more faithful to the provided knowledge and relevant to the dialogue history, as shown from the human evaluation on responses from GPT-2 and T5 models. In each section, * indicates that this result is significantly better ($p$-value $< 0.05$) from their respective baseline comparison. See Figure \ref{sec:he_detail} for a detailed visualization of the Likert score distribution.}
\label{tab:humaneval}
\end{table}

\subsection{Results with Retrieved Knowledge}
We reproduce the greedy decoding performance similar to what was reported in~\cite{zhao2020knowledge}. Here, we use the same scoring metrics on PICK as in \S\ref{sec:results_gold}. 
Although other decoding methods' underperform in comparison to greedy in the knowledge retrieval setting (see Table \ref{tab:knowledgpt}), PICK still shows improvement over each decoding method taken individually as the BLEU-4, ROUGE-L, F1, and KF1 scores all increase.
Interestingly, although our method improves the generated response's performance compared to its vanilla counterpart, the top performances are still comparable with the greedy decoding performance, except for the significantly better KF1 on the responses re-ranked by PICK. It is important to note that the KF1 here is calculated w.r.t the retrieved knowledge instead of the gold knowledge. We also investigate the underperformance of KnowledGPT being inferenced through decoding methods other than greedy, and we found that the issue persists through repeated trials in different settings. We leave this issue out of the scope of this paper.

\begin{figure}[!t]
\centering
\centering
  \includegraphics[width=\columnwidth]{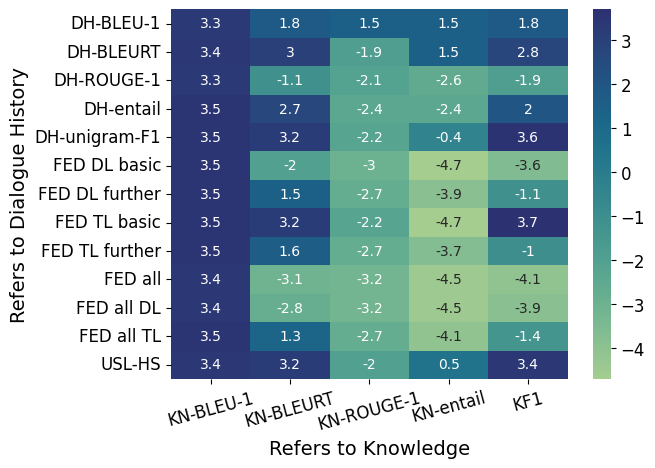}
\caption{Comparing combinations of automatic evaluation of response qualities w.r.t dialogue history and knowledge snippet, PICK with FED turn-level basic metrics and KF1 produced responses with the best qualities. The comparisons are shown here as a heatmap of the sum of mean-normalized BLEU-4, ROUGE-L, F1, and KF1 w.r.t gold responses. The x-axis labels the knowledge-oriented metrics, and the y-axis labels the dialogue-history-oriented metrics used in the comparisons. TL denotes turn level, and DL denotes dialogue level.}
\label{fig:ablation_selection_metrics}
\end{figure}

\subsection{Human Evaluation}

We conduct the manual evaluation on responses from GPT-2
and T5 models decoded with beam search 10. We compare both the vanilla (\textit{r}=1) and the PICK responses (\textit{r}=10) and evaluate the quality of the responses in the aspect of \textit{Relevance} and \textit{Faithfulness}.
Table \ref{tab:humaneval} shows that PICK responses from GPT-2 and T5 models are more faithful and relevant than the vanilla generations. These findings correlate well with the automatic results shown in Table \ref{tab:mainresults}, where for both of the responses, PICK responses achieved higher BLEU-4, ROUGE-L, F1, and KF1 scores in comparison to the vanilla responses. We attach a detailed visualization of the Likert score distribution in Figure \ref{sec:he_detail}.

It is also known from previous works that attempts to make the system more faithful usually lead to trade-offs between the response's relevance and faithfulness scores~\cite{rashkin2021increasing}, either due to the response being not quite as pertinent to the previous conversation turns or it gives overly extractive responses. This result also showcases the merit of the proposed re-ranking method to improve the faithfulness of the responses without a trade-off on the response's relevance towards the previous conversation turn.


\begin{figure*}[!t]
\centering
\begin{subfigure}{0.32\linewidth}
\centering
  \includegraphics[width=\linewidth]{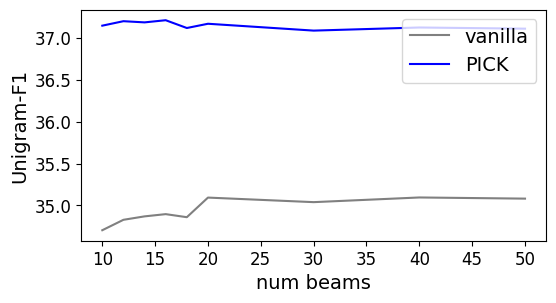}
\end{subfigure}
\begin{subfigure}{0.32\linewidth}
\centering
  \includegraphics[width=\linewidth]{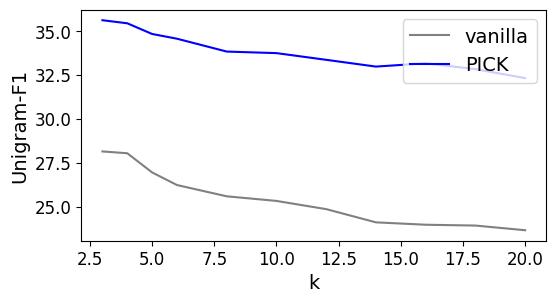}
\end{subfigure}
\begin{subfigure}{0.32\linewidth}
\centering
  \includegraphics[width=\linewidth]{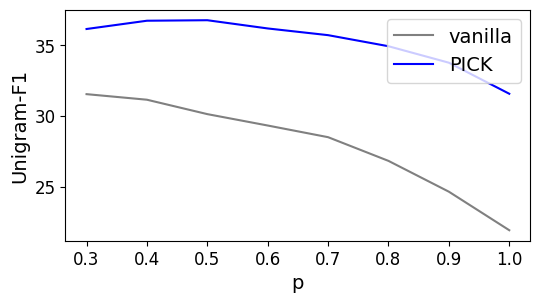}
\end{subfigure}
\caption{Performance comparison with unigram-F1 while varying the number of beams, $k$ and $p$ in respectively beam search, top-$k$ and top-$p$ sampling with $r$ kept at 10. Extending the number of beams does not help, but $k$ and $p$ threshold help to mitigate a bad response formation. 
}
\label{fig:ablation_decoding_method}
\end{figure*}

\begin{figure*}[!t]
\centering
\begin{subfigure}{0.32\linewidth}
\centering
  \includegraphics[width=\textwidth]{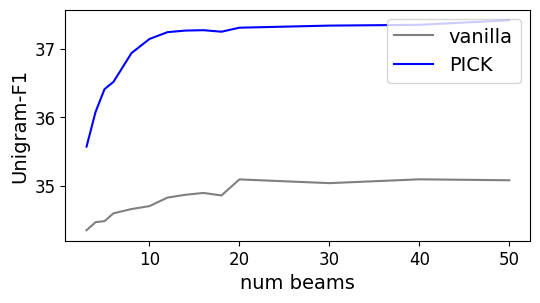}
\end{subfigure}
\hspace{15pt}
\begin{subfigure}{0.32\linewidth}
\centering
  \includegraphics[width=\textwidth]{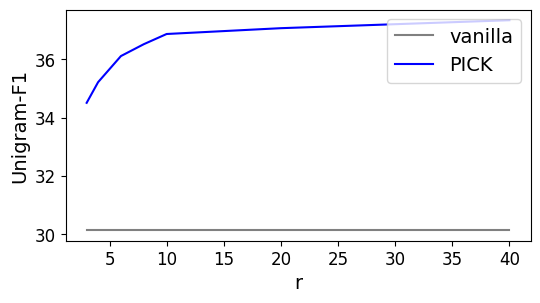}
\end{subfigure}
\caption{Performance comparison with unigram-F1 while varying $r$ to follow the number of beams (left figure) and varying $r$ on nucleus sampling with $p$ = 0.5 (right figure). Increasing $r$ in both the search and sampling decoding experiments promotes the existence of responses that are more similar to the gold response.
}
\label{fig:ablation_numretseq}
\end{figure*}

\section{Analysis and Discussion}

\subsection{Rescoring Metrics}
\label{sec:diff_metrics}

We study the effectiveness of automatic metrics explored in Section \ref{sec:faithfulness_score} and \ref{sec:relevance_score}. We employ GPT-2+PICK with each automatic metric as the scoring method by normalizing the performances on each of their mean and standard deviations for a fairer comparison. We report the normalized comparison as a heatmap in Figure \ref{fig:ablation_selection_metrics}. The best performance is achieved by using FED turn-level that measures the basic qualities of responses w.r.t the dialogue history (FED turn-level basic) alongside KF1.


\subsection{Decoding Strategy}

\label{sec:ablation_decoding}
We perform ablation of GPT-2+PICK and vanilla on the decoding methods to find the optimum configurations with our proposed method. For beam search, we increase the number of beams starting from $10$. For the top-$k$ and nucleus sampling, we increase the threshold of $k$ and $p$ to perform the sampling, starting from $k=3$ and $p=0.3$. For each experiment, we keep $r=10$, and we note the comparison of performances w.r.t the gold as a probe towards the degree of the responses being in a certain desired quality range that the gold responses reflect (Figure \ref{fig:ablation_decoding_method}). 


From Figure \ref{fig:ablation_decoding_method}, 
We observe that increasing the number of beams does not improve the performance of the generated responses, while as the $p$ and $k$ increase, the performances degrade. We conjecture that loosening the respective $p$ and $k$ sampling thresholds weakens the mitigation of bad continuations and, in turn, produces worse generations (i.e., generations with considerably low perplexity \cite{holtzman2019curious}). The two sampling strategies show different optimum thresholds, as top-$k$ sampling responses start to deteriorate with $k$ larger than $3$, while for the nucleus sampling, we see benefits in relaxing the $p$ threshold to $0.5$. We conjecture that due to the probability selection of the nucleus sampling, top-$p$ tokens within $p\leq0.5$ could still be reliable, as it produces generations with considerably low perplexity as mentioned in~\cite{holtzman2019curious}.

\subsection{Number of Return Sequences (\textit{r})}
\label{sec:ablation_numretseq}

We also perform further ablation of GPT-2+PICK and vanilla on varying $r$. We extend $r$ on the nucleus sampling with $p=0.5$, and on the beam search, we set the $r$ to follow the number of beams $n$ to retain the top $n$ choices when a new token in the sequence is generated.
The performance comparisons are noted in Figure \ref{fig:ablation_numretseq}.


Our observation shows that increasing $r$ in both the search and sampling decoding experiments promotes the existence of responses that are more similar to the gold response. With the gold response holding the desired qualities we aim to achieve, these findings also indicate that increasing the $r$ could help increase the response qualities generated by the same model to some extent. Figure \ref{fig:ablation_numretseq}. shows that PICK response performance begins to saturate around $n=10$ and $r=10$ in the beam search and nucleus sampling experiment, most likely because the best candidate response is consistently found within that range of return sequences.


\subsection{Error Analysis}

To better understand our method's limitation, we provide manually sampled study cases with GPT-2+PICK using beam search (\textit{n} = 50, \textit{r} = 50), in which better responses are not selected by the scorer. We observe three kinds of errors. First, the current metric fails to promote the selection of better responses. We conjecture that this happens due to the low correlation of the automatic metrics towards the human judgments \cite{yeh2021comprehensive}; hence, implementing better human preference metrics will aid in better response promotion.

Second, in some cases, substandard responses are selected due to their high overlap with the knowledge snippet. This could be because the metrics used rely on the spurious correlation between attribution and word overlap and thus do not reliably distinguish attributable abstractive responses \cite{mccoy2019right, dziri2022evaluating}. We perform a further ablation study on this error in Appendix \ref{sec:kn_copy}. Third, the knowledge snippets provided are irrelevant to the dialogue history.
We provide these entries of case study samples in Table \ref{tab:error_samples}.

\section{Conclusion}

This work investigates the alignment of KGD responses to faithfulness and relevance. We propose PICK, a straightforward yet effective generation re-ranking framework for KGD. PICK is model-agnostic, does not require further model tuning, nor requires additional labelled data for the language modelling alignment. Experimental results show that the proposed method enables models to produce better responses that are more faithful to the provided knowledge and relevant to the dialogue history.

\section*{Acknowledgements}

We thank the anonymous reviewers for their valuable and constructive comments. We thank Tiezheng Yu for the insightful discussions. This work has been partially funded by the PF20-43679 Hong Kong PhD Fellowship Scheme, Research Grant Council, Hong Kong, and the Hong Kong Fellowship Scheme by the Hong Kong Research Grants Council (RGC).

\section*{Limitations}

While our proposed method, PICK, is model-agnostic and can be adopted by various model architectures, our exploration in this work is limited to GPT-2 and T5. Generating multiple alternative responses within a single decoding process also can increase the computational overhead of the KGD system, thus ways to increase PICK's efficiency would be beneficial in the future. Additionally, while PICK improves the faithfulness and relevance of responses, it may not address other challenges in knowledge-grounded dialogue systems such as self-consistency, engagingness, long-term coherence, and more. Further research is needed to explore these limitations and develop more comprehensive approaches for generating better responses. We leave these explorations open for future work.

\section*{Ethics Statement}

In this paper, we propose a re-ranking framework, targeting better correlating the final generation with some concrete attributes of the response. However, our work has a broader impact given the current popularity of ChatGPT. ChatGPT replies on a reward model to model human feedback for reinforcement learning. However, the training of the reward model requires a huge amount of human annotations, which is time- and resource-consuming. Then it comes to a question -what is the expression of human preference and whether it is possible to model human preference without heavy human annotations? Though far from perfect, we take an initial step in this direction by exploring the usage of automatic metrics to re-rank the responses. We believe it is a promising and valuable research topic.

\bibliography{custom}
\bibliographystyle{acl_natbib}

\clearpage
\appendix
\onecolumn

\setcounter{table}{0}
\renewcommand{\thetable}{A\arabic{table}}
\setcounter{figure}{0}
\renewcommand{\thefigure}{A\arabic{figure}}

\section{Dialogue samples}
\label{sec:dialogue-samples}

We provide the dialogue samples in Table~\ref{tab:cherrypicked_samples}.

\begin{table*}[ht]
\centering
\resizebox{\textwidth}{!}{%
\begin{tabular}{ll}
\toprule
\multicolumn{2}{l}{\textbf{Knowledge snippet}} \\
\multicolumn{2}{l}{Due to his powerful and very large vocal range and energetic live performances, Rose has been named one of the greatest singers of all time by various} \\ 
\multicolumn{2}{l}{media outlets, including "Rolling Stone" and "NME".} \\
\multicolumn{2}{l}{\textbf{Dialogue History}} \\
\multicolumn{2}{l}{Speaker 1: Didn't their guitarist slash leave the band?} \\
\multicolumn{2}{l}{Speaker 2: When did he leave did he release the six albums with them} \\
\multicolumn{2}{l}{Speaker 1: I heard Axl Rose was known for throwing tantrums} \\
\multicolumn{2}{l}{\textbf{Response}} \\
Vanilla & He was known for throwing tantrums \\
PICK & He has been named one of the greatest singers of all time by various media outlets \\
Gold & Maybe that was due to his powerful and large vocal range \\ \midrule
\multicolumn{2}{l}{\textbf{Knowledge snippet}} \\
\multicolumn{2}{l}{His music career began there in 1954, recording at Sun Records with producer Sam Phillips, who wanted to bring the sound of African American music} \\
\multicolumn{2}{l}{to a wider audience.} \\
\multicolumn{2}{l}{\textbf{Dialogue History}} \\
\multicolumn{2}{l}{Speaker 1: I like to listen to Elvis! Have you heard him?} \\
\multicolumn{2}{l}{Speaker 2: Yeah my favorite song of his is "Hound Dog" written by Jerry Leiber.} \\
\multicolumn{2}{l}{Speaker 1: Yes, I love that song! I really like his version of "My Way"} \\
\multicolumn{2}{l}{\textbf{Response}} \\
Vanilla & He started his career in 1954, and recorded his first album in the early 60s. \\
PICK & His music career began in 1954, he was a producer and he wanted to bring the sound of African American music to a wider audience. \\
Gold & Elvis started his music career in 1954 recording at Sun Records with producer Sam Phillips. \\ \midrule
\multicolumn{2}{l}{\textbf{Knowledge snippet}} \\
\multicolumn{2}{l}{, the books have sold more than 500 million copies worldwide, making them the best-selling book series in history, and have been translated into} \\
\multicolumn{2}{l}{seventy-three languages.} \\
\multicolumn{2}{l}{\textbf{Dialogue History}} \\
\multicolumn{2}{l}{Speaker 1: I liked it because of all the different animals, I am in love with animals, and the ones they made were so unique!} \\
\multicolumn{2}{l}{Speaker 2: Harry Potter has attracted a wide adult audience, as well as younger readers. If you go to Universal you will see whole families dressed up!} \\
\multicolumn{2}{l}{Speaker 1: I know everyone was young when everyone was into it, it was huge when I was young, like all the rage} \\
\multicolumn{2}{l}{\textbf{Response}} \\
Vanilla & It's hard to imagine how much the books have sold worldwide!It must be so popular now! \\
PICK & I agree, the books have sold more than 500 million copies worldwide, and they have been translated into seventy-three languages. \\
Gold & The books have sold 500 million copies worldwide, making it the best-selling book series in history. Wow that is impressive! \\ \bottomrule
\end{tabular}%
}
\caption{Samples of the responses from the beam search (n=10), top-$k$ sampling, and nucleus sampling generated by GPT-2 using the vanilla, re-ranked, and oracle methods. We also attach the gold responses for each dialogue. The vanilla response is not as faithful w.r.t. the knowledge snippet, especially compared to the other 3 responses. The re-ranked responses, however, are more faithful and yet still relevant to the dialogue history as they are not a verbatim copy of the knowledge snippet.}
\label{tab:cherrypicked_samples}
\end{table*}

\section{Details of Human Evaluation Results}
\label{sec:he_detail}

\begin{figure}[htbp]
\centering
\centering
  \includegraphics[width=0.9\columnwidth]{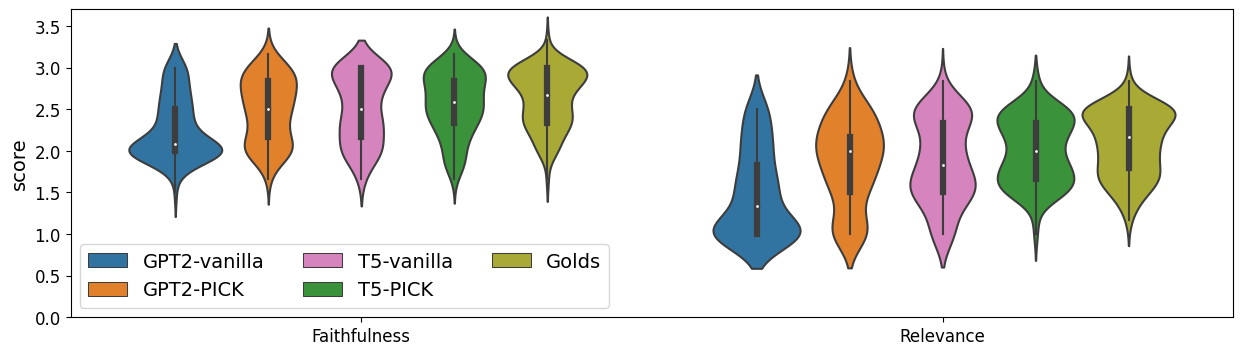}
\caption{Distribution of human evaluation's Likert score on Faithfulness and Relevance}
\label{fig:distribution_he}
\end{figure}

We further show the distribution of Likert scores in Figure~\ref{fig:distribution_he} in addition to the average reports.

\section{Error analysis}
\label{sec:error_analysis}

We provide the dialogue samples of our error analysis in Table~\ref{tab:error_samples}.

\begin{table*}[ht]
\centering
\resizebox{\textwidth}{!}{%
\begin{tabular}{ll}
\toprule

\multicolumn{2}{l}{\textbf{Knowledge snippet}} \\
\multicolumn{2}{l}{The album has sold approximately 30 million copies worldwide, including 18 million units in the United States, making it the best-selling debut album of all time in} \\ 
\multicolumn{2}{l}{the US, as well as the eleventh best-selling album in the United States.} \\ 
\multicolumn{2}{l}{\textbf{Dialogue History}} \\
\multicolumn{2}{l}{Speaker 1: I totally forgot about that song! Didn't listen to it for a while!} \\
\multicolumn{2}{l}{Speaker 2: You know that album sold 30 million copies!} \\
\multicolumn{2}{l}{Speaker 1: Really? That's insane! I didn't know that} \\
\multicolumn{2}{l}{\textbf{Response}} \\
PICK & The album has sold approximately 30 million copies worldwide, including 18 million units in the United States, making it the best-selling debut album\\
Other candidate & It is the best selling debut album of all time! \\
Gold & Maybe that was due to his powerful and large vocal range \\ \midrule

\multicolumn{2}{l}{\textbf{Knowledge snippet}} \\
\multicolumn{2}{l}{Thierry Daniel Henry (; born 17 August 1977) is a retired French professional footballer who played as a forward and is the second assistant manager of the Belgium } \\ 
\multicolumn{2}{l}{national team.} \\ 
\multicolumn{2}{l}{\textbf{Dialogue History}} \\
\multicolumn{2}{l}{Speaker 1: No i didn't know that! that's pretty cool!} \\
\multicolumn{2}{l}{Speaker 2: Yes, he actually played for Monaco, Barcelona, New York, and Arsenal, where he was the club's all-time record goalscorer!} \\
\multicolumn{2}{l}{Speaker 1: Wow! I bet he is really good at what he is doing. How old is he?} \\
\multicolumn{2}{l}{\textbf{Response}} \\
PICK & Thierry Henry is a retired French professional footballer who played as a forward and is the second assistant manager of the Belgium national team\\
Other candidate & He was born on August 17, 1977. \\
Gold & He was born on 17 August 1977 which makes him 40 years old. Do you like watching soccer? \\ \bottomrule

\multicolumn{2}{l}{\textbf{Knowledge snippet}} \\
\multicolumn{2}{l}{Dylan Lauren (born May 9, 1974) is an American entrepreneur.} \\ 
\multicolumn{2}{l}{\textbf{Dialogue History}} \\
\multicolumn{2}{l}{Speaker 1: I love chocolate bars. Have you ever heard of Dylan's Candy Bar?} \\
\multicolumn{2}{l}{Speaker 2: I have heard of it, the daughter of Dylan Lauren owns it, but she is a fashion designer} \\
\multicolumn{2}{l}{Speaker 1:  Who is Dylan Lauren?} \\
\multicolumn{2}{l}{\textbf{Response}} \\
PICK & Dylan Lauren (born May 9, 1974) is an American entrepreneur.\\
Other candidate & She is an american entrepreneur! \\
Gold & She is an american entrepreneur! \\ \midrule

\multicolumn{2}{l}{\textbf{Knowledge snippet}} \\
\multicolumn{2}{l}{The first match of American football was played on November 6, 1869, between two college teams, Rutgers and Princeton, under rules } \\ 
\multicolumn{2}{l}{based on the association football rules of the time.} \\ 
\multicolumn{2}{l}{\textbf{Dialogue History}} \\
\multicolumn{2}{l}{Speaker 1: Same here. My favorite team is the Saints!} \\
\multicolumn{2}{l}{Speaker 2: The first football game was played on November 6, 1869.} \\
\multicolumn{2}{l}{Speaker 1: Wow! I did not know it was that old. Do you know who the teams were?} \\
\multicolumn{2}{l}{\textbf{Response}} \\
PICK & The first match of American football was played on November 6, 1869 between two college teams, Rutgers and Princeton, under rules\\
& based on the association\\
& football rules of the time.\\
Other candidate & The teams were Rutgers and Princeton. \\
Gold & The teams were Rutgers and Princeton. Two college teams. \\ \bottomrule

\multicolumn{2}{l}{\textbf{Knowledge snippet}} \\
\multicolumn{2}{l}{The maximum score is 300, which is achieved by getting 12 strikes in a row.} \\  
\multicolumn{2}{l}{\textbf{Dialogue History}} \\
\multicolumn{2}{l}{Speaker 1: That's cool! So I can just get started again. :) Do you know much about bowling?} \\
\multicolumn{2}{l}{Speaker 2: I am knowledgeable. In bowling your target is to knock all pins in the lane. } \\
\multicolumn{2}{l}{Speaker 1: Oh yeah? What happens if you don't hit any pins?} \\
\multicolumn{2}{l}{\textbf{Response}} \\
PICK & The maximum score is 300 which is achieved by getting 12 strikes in a row\\
Other candidate & If you don't hit any pins, you get 12 strikes in a row. \\
Gold & If you will make a strike when you don't hit any pins. \\ \midrule

\multicolumn{2}{l}{\textbf{Knowledge snippet}} \\
\multicolumn{2}{l}{To Kill a Mockingbird is a novel by Harper Lee published in 1960.} \\  
\multicolumn{2}{l}{\textbf{Dialogue History}} \\
\multicolumn{2}{l}{Speaker 1: Oh wow, a Jack of all Trades it seems. Do you have a favorite novel? I'd love a recommendation!} \\
\multicolumn{2}{l}{Speaker 2: His first, "A Time to Kill" is older, it was published in 1989 but he spent four years writing it and in my opinion, it's his best.} \\
\multicolumn{2}{l}{Speaker 1: 4 years is an incredible amount of time to spend on a book, I bet it is good. I will definitely check it out! I imagine it's a thriller about murder?} \\
\multicolumn{2}{l}{\textbf{Response}} \\
PICK & It was published in 1960 by Harper Lee.\\
Other candidate & I'm not sure, but I do know that to Kill a Mockingbird is a book by Harper Lee. \\
Gold & Yes, another book I like is To Kill a Mockingbird but that was written by Harper Lee. That is also about a crime \\ \bottomrule

\end{tabular}%
}
\caption{Six study case samples from the GPT-2+PICK using beam search (\textit{n} = 50, \textit{r} = 50). The first two samples show errors caused by the current metric failing to promote the selection of better responses seen in the other candidate. The next two samples show errors caused by the selection of worse responses due to high overlap towards the knowledge snippet. The last two sample errors were because of the utilization of irrelevant provided knowledge snippets. The other candidate is one of the best response alternatives that are not selected either by vanilla or even by PICK.}
\label{tab:error_samples}
\end{table*}





\clearpage


\section{Re-ranking responses w.r.t the dialogue history, the knowledge, or both}
\label{sec:kn_copy}

We also perform an ablation study of doing the re-ranking with different considerations on the scoring method: only considering the response quality w.r.t the dialogue history (using FED turn-level basic metrics), the knowledge (using unigram-F1 to the knowledge), or both (see Table \ref{tab:selectionreferals}). Re-ranking the response candidates considering only the qualities w.r.t the dialogue history results in generations that are less knowledgeable and perform worse across the board. 
 On the contrary, considering only how much of the response refers to the knowledge snippets promotes more responses that are direct copies of the knowledge. While considering both the response candidate's qualities w.r.t the dialogue history and the knowledge snippet produces similar response quality per their automatic metrics performances, the responses contain far less direct knowledge copies within them. This is a desirable result since the direct knowledge copies responses could be considered responses less relevant to the dialogue as they would be perceived as very rigid non-human-like responses. This is also an interesting finding, that the combination of both qualities in consideration can alleviate the failure of overlap-based metrics to rank abstractive but faithful responses as previously observed in \cite{dziri2022evaluating}.

\begin{table*}[ht]
\centering
\resizebox{\textwidth}{!}{%
\begin{tabular}{lcccccccccc}
\toprule
\multirow{2}{*}{\textbf{Models}} & \multicolumn{5}{c}{\textbf{Test (seen topics)}} & \multicolumn{5}{c}{\textbf{Test (unseen topics)}} \\ \cmidrule(lr){2-6} \cmidrule(lr){7-11}
 & \textbf{BLEU-4} & \textbf{ROUGE-L} & \textbf{F1} & \textbf{KF1} & \textbf{\% Kn-copy} & \textbf{BLEU-4} & \textbf{ROUGE-L} & \textbf{F1} & \textbf{KF1} & \textbf{\% Kn-copy} \\ \midrule
Beam search & 15.4 & 33.4 & 35.8\% & 68.9\% & 24.3\% & 14.3 & 32.5 & 34.7\% & 64.5\% & 15.3\% \\
\quad + PICK w.r.t dialogue history & 10.9 & 31.7 & 33.7\% & 54.7\% & \textbf{0.8\%} & 10.4 & 30.9 & 32.8\% & 52.2\% & \textbf{0.6\%} \\
\quad + PICK w.r.t knowledge & \textbf{16.9} & \textbf{34.5} & \textbf{37.4\%} & \textbf{81.1\%} & 36.1\% & \textbf{16.0} & 34.4 & \textbf{37.1\%} & \textbf{78.8\%} & 26.6\% \\
\quad + PICK w.r.t both & 16.7 & \textbf{34.5} & 37.4\% & 80.4\% & 15.5\% & \textbf{16.0} & \textbf{34.5} & \textbf{37.1\%} & 78.2\% & 13.7\% \\ \bottomrule
\end{tabular}%
}
\caption{Ablation on PICK w.r.t dialogue history, knowledge, or both. PICK that considers both references produce similar performances but with responses containing less direct knowledge copies within them (i.e. 15.5\% vs 36.1\%) compared to only conditioning it to the knowledge. We use beam search with ($n$ = 10, $r$ = 10). The best performances in each section are in \textbf{bold}.}
\label{tab:selectionreferals}
\end{table*}


\section{Annotation guideline}
\label{sec:appendix_annotation guidance}

We gave the following annotation instructions, guidelines, and examples to the human evaluator for them to follow.

\begin{figure*}[!t]
\centering
\begin{subfigure}{\linewidth}
\centering
  \includegraphics[width=\linewidth]{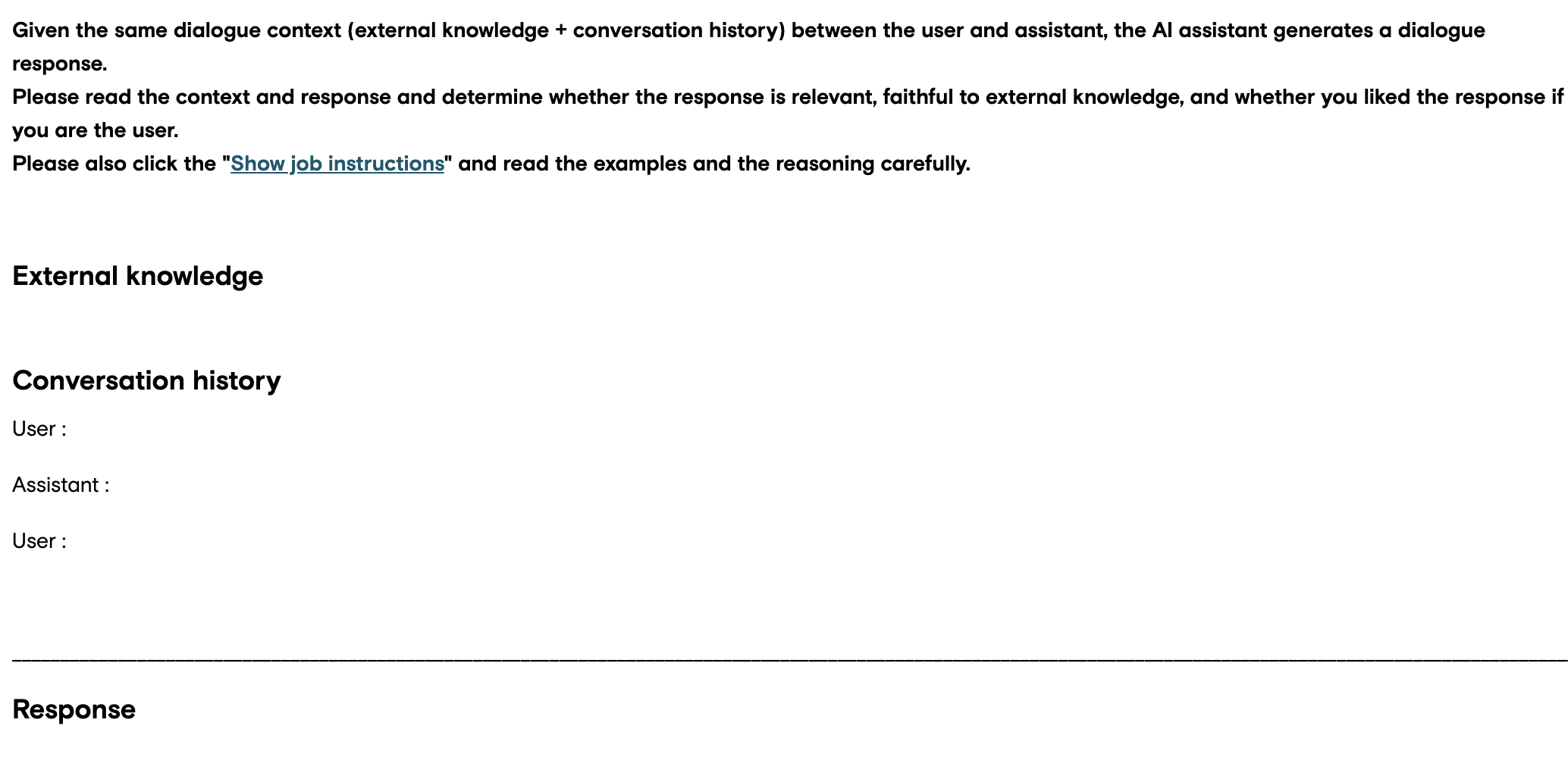}
  \caption{}
\end{subfigure}
\begin{subfigure}{\linewidth}
\centering
  \includegraphics[width=\linewidth]{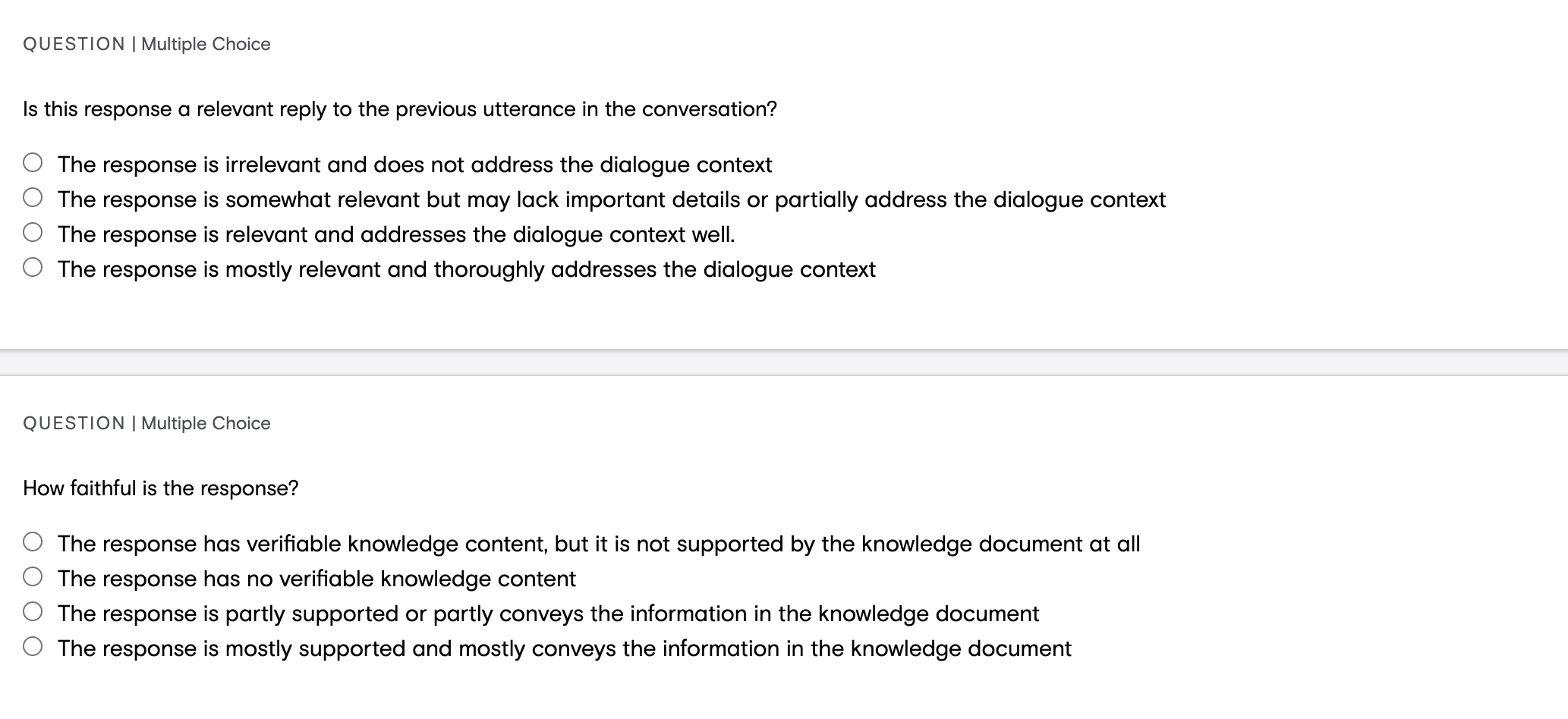}
  \caption{}
\end{subfigure}
\caption{Main instruction (a) and annotation guidelines (b) for the human evaluation.
}
\label{fig:annotation_guideline_main}
\end{figure*}

\begin{figure*}[!t]
\centering
\begin{subfigure}{\linewidth}
\centering
  \includegraphics[width=\linewidth]{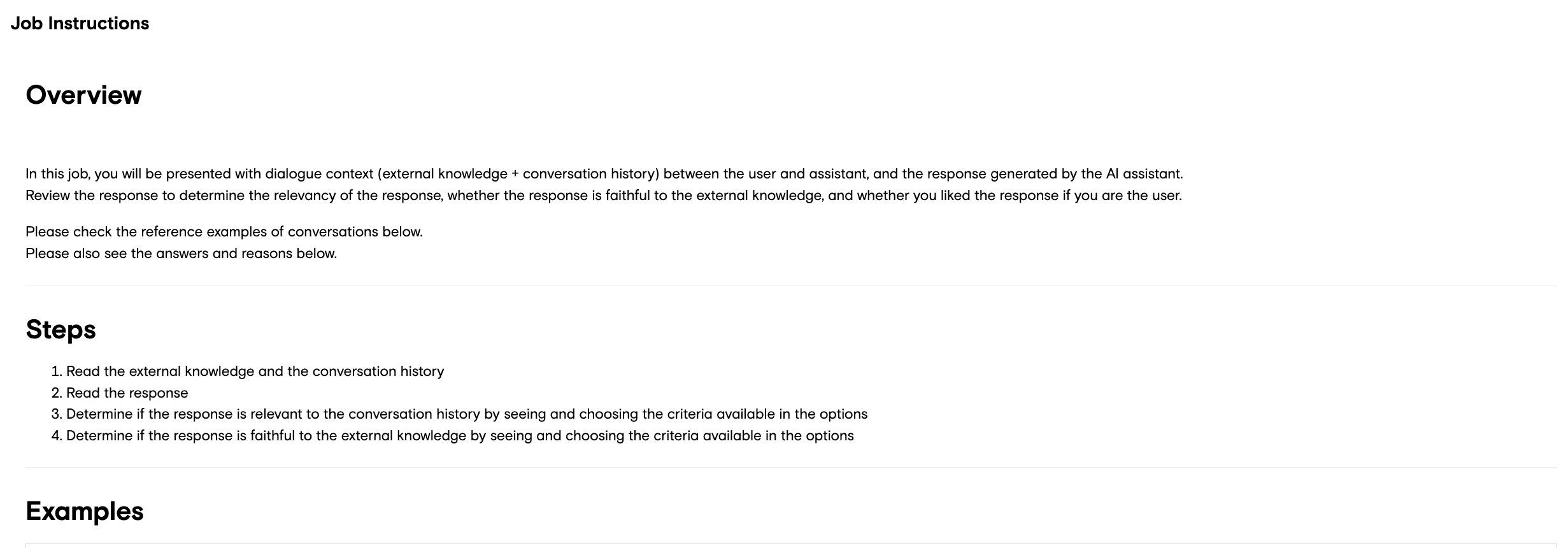}
  \caption{}
\end{subfigure}
\begin{subfigure}{\linewidth}
\centering
  \includegraphics[width=\linewidth]{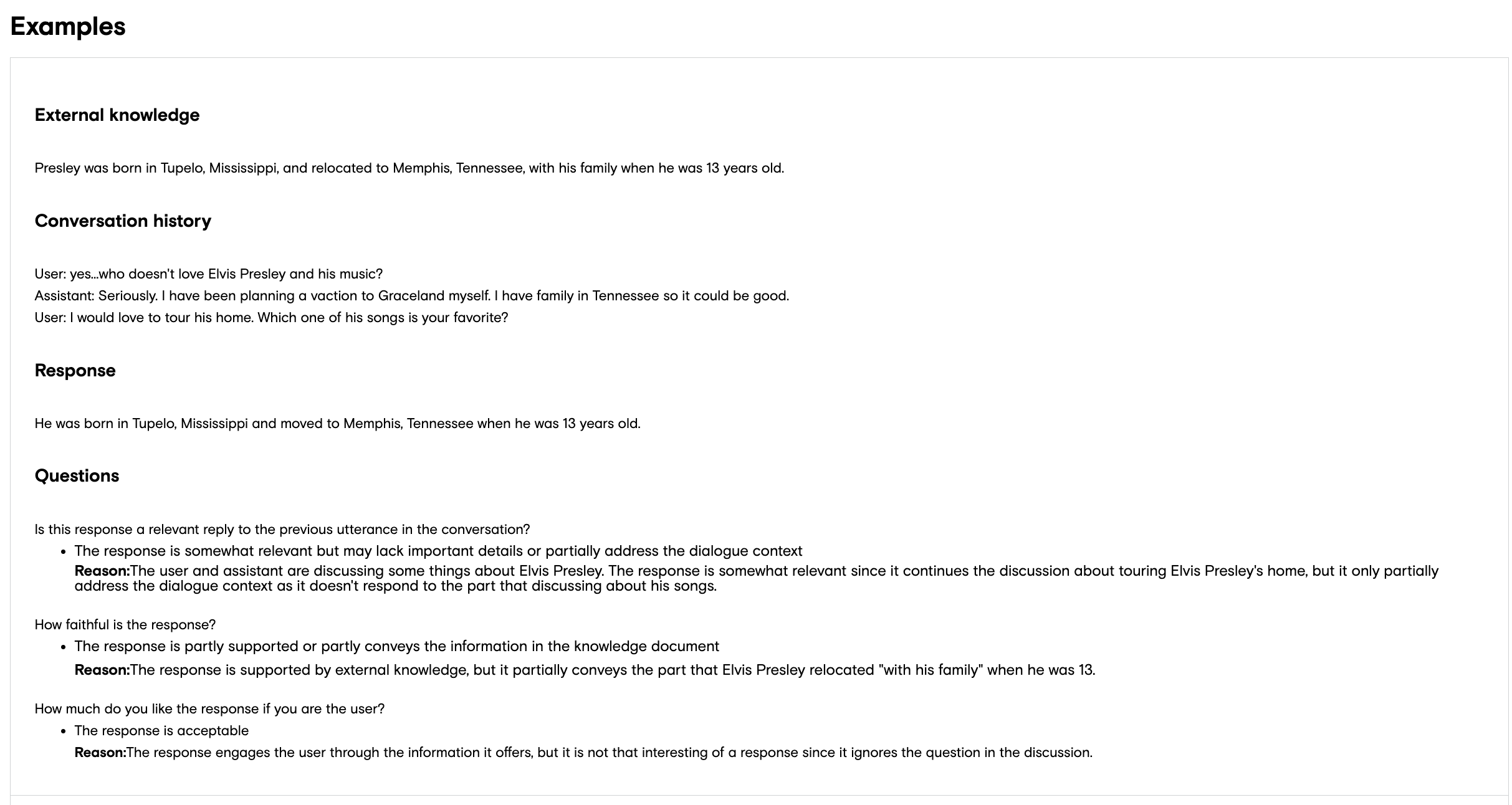}
  \caption{}
\end{subfigure}
\caption{Additional instructions of human evaluation (a) and the examples of annotation (b).
}
\label{fig:annotation_guideline_add}
\end{figure*}

\end{document}